\documentclass[10pt,twocolumn,letterpaper]{article}

\usepackage{cvpr}
\usepackage{times}
\usepackage{epsfig}
\usepackage{graphicx}
\usepackage{amsmath}
\usepackage{amssymb}
\usepackage{multirow}
\usepackage{booktabs}
\usepackage[table]{xcolor}
\usepackage{threeparttable}
\usepackage{colortbl}
\usepackage{algorithm}
\usepackage{algorithmic}

\definecolor{mygray}{gray}{.92}
\usepackage{bm}

\DeclareMathOperator{\Tr}{Tr}

\DeclareMathOperator*{\argmax}{argmax}

\makeatletter
\newcommand{\thickhline}{%
    \noalign {\ifnum 0=`}\fi \hrule height 1pt
    \futurelet \reserved@a \@xhline
}
\makeatother

\usepackage[pagebackref=true,breaklinks=true,letterpaper=true,colorlinks,bookmarks=false]{hyperref}

\cvprfinalcopy 

\def\cvprPaperID{3909} 

\ifcvprfinal\fi
\begin{document}

\title{Reasoning Visual Dialogs with Structural and Partial Observations}

\author{Zilong Zheng\thanks{Equal contribution.} $~^{1}$, Wenguan Wang\footnotemark[1] $~^{2,1}$, Siyuan Qi\footnotemark[1] $~^{1,3}$, Song-Chun Zhu$~^{1,3}$\\
$^1$\small University of California, Los Angeles, USA~~~~$^2$\small Inception Institute of Artificial Intelligence, UAE\\
$^3$\small International Center for AI and Robot Autonomy (CARA)\\
{\tt\small zilongzheng0318@ucla.edu, wenguanwang.ai@gmail.com, syqi@cs.ucla.edu, sczhu@stat.ucla.edu}\\
{\tt\small \url{https://github.com/zilongzheng/visdial-gnn}}
}

\maketitle

\thispagestyle{empty}

\begin{abstract}
We propose a novel model to address the task of Visual Dialog which exhibits complex dialog structures. To obtain a reasonable answer based on the current question and the dialog history, the underlying semantic dependencies between dialog entities are essential.
In this paper, we explicitly formalize this task as inference in a graphical model with partially observed nodes and unknown graph structures (relations in dialog). The given dialog entities are viewed as the observed nodes. The answer to a given question is represented by a node with missing value.
We first introduce an Expectation Maximization algorithm to infer both the underlying dialog structures and the missing node values (desired answers). Based on this, we proceed to propose a differentiable graph neural network (GNN) solution that approximates this process.
Experiment results on the VisDial and VisDial-Q datasets show that our model outperforms comparative methods. It is also observed that our method can infer the underlying dialog structure for better dialog reasoning.
\end{abstract}

\vspace{-3pt}
\section{Introduction}
\begin{figure}[t]
  \centering
      \includegraphics[width=1 \linewidth]{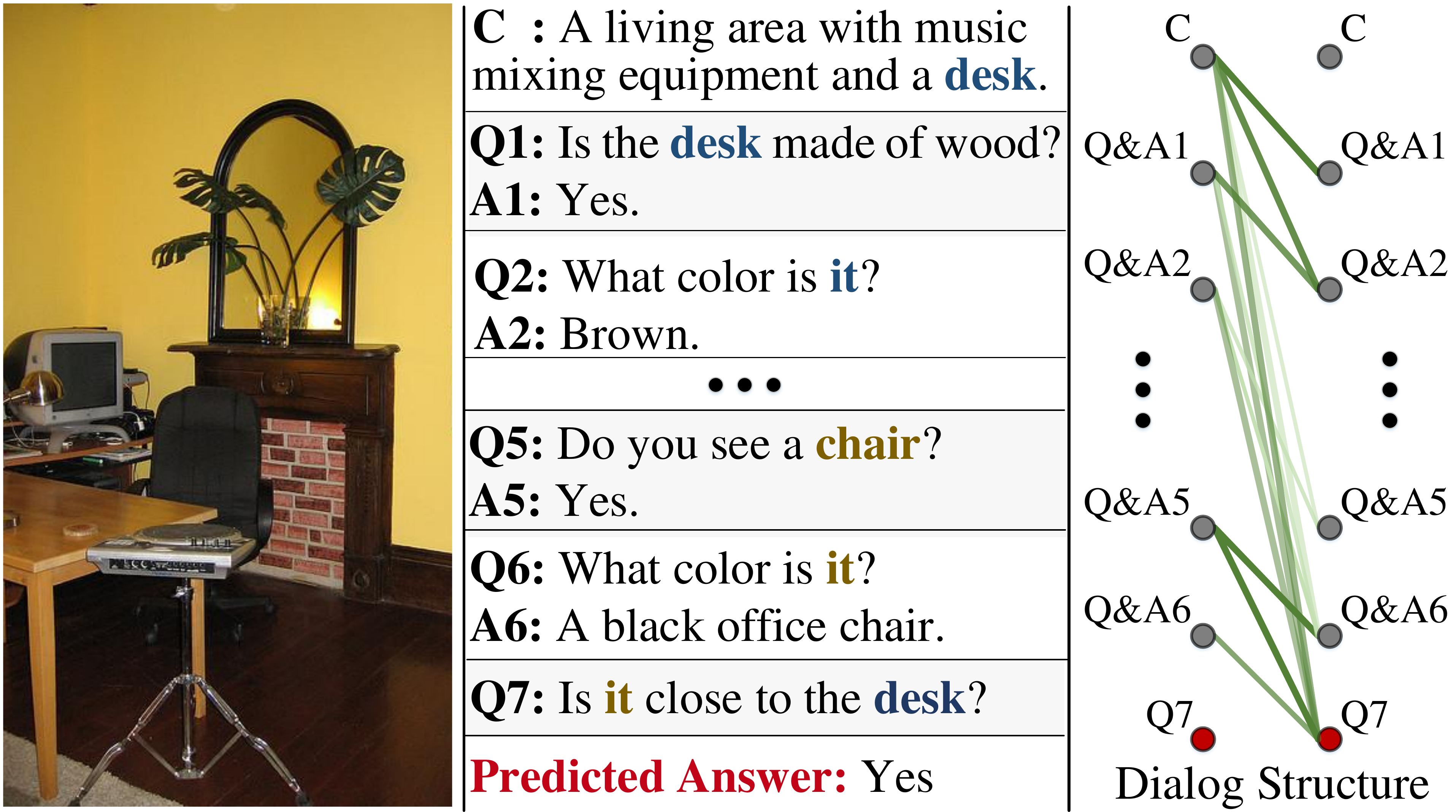}
\caption{\small An illustration of the visual dialog task. Left: context image. Middle: image caption, dialog history, current query question, and the predicted answer. Right: the underlying semantic dependencies between nodes in the dialog (darker green links indicate higher dependencies). }
\label{fig:idea}
\vspace{-12pt}
\end{figure}

\vspace{-3pt}
Visual Dialog has drawn increasing research interests
at the intersection of computer vision and natural language
processing. In such tasks, an image is given as context input, associated with a summarizing caption and a dialog history of question-answer pairs. The goal is to answer questions posed in natural language about images~\cite{das2017visual}, or recover a follow-up question based on the dialog history~\cite{jain2018two}.
Despite its significance to artificial intelligence and human-computer interaction,  it poses a richer set of challenges (see an example in \autoref{fig:idea}) -- requiring representing/understanding a series of multi-modal entities 
, and reasoning the rich semantic relations/structures among them. An ideal inference algorithm should be able to find out the underlying semantic structure and give a reasonable answer based on this structure.

Previous studies have explored this task through embedding rich features from image representation learned from convolutional neural networks and language (\ie, question-answer pairs, caption) representations
learned from recurrent sequential models. Their impressive results well demonstrate the importance of mining and fusing multi-modal information in this area. However, they largely neglect the key role of the rich relational information in dialog. Although a few~\cite{zhuang2018parallel,Wu_2018_CVPR} leveraged co-attention mechanisms to capture cross-modal correlations, 
their reasoning ability is still quite limited. They typically concatenate the multi-modal features together and directly project the concatenated feature into the answer feature space by a neural network. On one hand, their reasoning process does not fully utilize the rich relational information in this task due to their monolithic vector representations of dialog. On the other hand, their feed-forward network designs fail to deeply and iteratively mine and reason the information from different dialog entities over the inherent dialog structures.

In this work, we consider the problem of recovering the dialog structure and reasoning about the question/answer simultaneously. We represent the dialog as a graph, where the nodes are dialog entities and the edges are semantic dependencies between nodes. Given the dialog history as input, we have a partial observation of the graph. Then we formulate the problem as inferring about the values of unobserved nodes (\eg, the queried answer) and the graph structure.

The challenge of the problem is that there is no label for dialog structures. For each individual dialog, we need to recover the underlying structure in an unsupervised manner. The node values could then be inferred iteratively with the graph structure: we can reason about the nodes based on the graph structure, and further improve the structure based on the node values.
To tackle this challenge, the insight is that a graph structure essentially specifies a joint probability distribution for all the nodes in the graph. Therefore we can view the queried dialog entities as missing values in the data, the dialog structure as unknown parameters of the distribution. Specifically, we encode the dialog as a Markov Random Field (MRF) where some nodes are observed, and the goal is to infer the edge weights between nodes as well as the value of unobserved nodes. We formulate a solution based on the Expectation-Maximization (EM) algorithm, and provide a graph neural network (GNN) approach to approximate this inference.

%

Our model provides a unified framework which is applicable to diverse dialog settings (detailed in \autoref{sec:ex}). Besides, it provides extra post-hoc interpretability to show the dialog structures through an implicit learning manner.
We evaluate the performance of our method on VisDial v0.9~\cite{das2017visual}, VisDial v1.0~\cite{das2017visual} and VisDial-Q~\cite{jain2018two} datasets.
The experimental results prove that our model is able to automatically parse the dialog structure and infer reasonable answer, and further achieves promising performance.


%
\vspace{-3pt}
\section{Related Work}\label{sec:rw}
\vspace{-3pt}


\noindent\textbf{Image Captioning} aims to annotate images
with natural language at the scene level automatically, which has been a long-term active research area in computer vision community. Early work~\cite{ordonez2011im2text,gong2014improving} typically tackled this task as a retrieval problem, \ie, finding the best fitting caption from a set of predetermined caption templates. Modern methods~\cite{mao2014deep,karpathy2015deep,vinyals2015show} were mainly based on a CNN-RNN framework, where the RNN leverages the CNN-representation of an input image to output a word sequence as the caption. In this way, they were freed from dependence of the pre-defined, expression-limited caption candidate pool. After that, some methods~\cite{xu2015show,Anderson_2018_CVPR,lu2017knowing} tried to integrate the vanilla CNN-RNN architecture with neural attention mechanisms, like semantic attention~\cite{lu2017knowing}, and bottom-up/top-down attention~\cite{Anderson_2018_CVPR}, to name a few representative ones.  Another popular trend~\cite{gan2017,park2017attend,johnson2016densecap,Chen_2018_CVPR,Mathews_2018_CVPR,Luo_2018_CVPR,Chen_2018_ECCV} in this area focuses on improving the discriminability of caption generations, such as stylized image captioning~\cite{gan2017,Chen_2018_ECCV}, personalized image captioning~\cite{park2017attend}, and context-aware image captioning~\cite{johnson2016densecap,Chen_2018_CVPR}.

\noindent\textbf{Visual Question Answering} focuses on providing a natural language answer
given an image and a free-form, open-ended question. It is a more recent (dated back to~\cite{malinowski2014multi,Antol_2015_ICCV}) and challenging task (need to access information from both the question and image). With the availability of large-scale datasets~\cite{ren2015exploring,Antol_2015_ICCV,gao2015you,goyal2017making,johnson2017clevr}, numerous VQA models were proposed to build multimodal representations upon the CNN-RNN architecture~\cite{gao2015you,ren2015exploring}, and recently extended with differentiable attentions~\cite{xu2015show,lu2016hierarchical,yang2016stacked,zhu2016visual7w,Anderson_2018_CVPR,Malinowski_2018_ECCV}. Rather than above classification-based VQA models, there were some other work~\cite{shih2016look,jabri2016revisiting,Teney_2018_CVPR,bai2018deep} leveraged answer representations into the VQA reasoning, \ie, predicting whether or not an image-question-answer triplet is correct. Teney~\etal~\cite{teney2017graph} proposed to solve VQA with graph-structured representations of both visual content and questions, showing the advantages of graph neural network in such structure-rich problems. Narasimhan~\etal~\cite{narasimhan2018out} applied graph convolution networks for factual VQA.  However, there are some notable differences between our model and~\cite{teney2017graph,narasimhan2018out} in the fundamental idea and theoretical basis, besides the specific tasks. First, we model the visual dialog task as a problem of inference over a graph with partially observed data and unknown dialog structures. This is one step further than propagating information over a fixed graph structure. Second, we emphasize both graph structure inference (in a unsupervised manner) and unobserved node reasoning. Last, the proposed model provides an end-to-end network architecture to approximate the EM solution and offers a new glimpse into the visual dialog task.

\noindent\textbf{Visual Dialog} refers to the task of answering a sequence
of questions about an input image~\cite{das2017visual,de2017guesswhat}. It is the latest vision-and-language problem, after the popularity of image captioning and visual question answering. It requires to reason about the image, the on-going question, as well as the past dialog history. \cite{das2017visual} and~\cite{de2017guesswhat} represented two early attempts towards this direction, but with different dialog settings. In~\cite{das2017visual}, a VisDial dataset is proposed and the questions in this dataset are free-form and may concern arbitrary content of the images. Differently, in~\cite{de2017guesswhat}, a `Guess-What' game is designed to identify a secret object through a series of yes/no questions. Following~\cite{das2017visual}, Lu \etal~\cite{lu2017best} introduced a generator-discriminator architecture, where the generator are improved using a perceptual loss from the pre-trained discriminator. In~\cite{seo2017visual}, a neural attention mechanism, called Attention Memory,  is proposed to resolve the current reference in the dialog. Das \etal~\cite{das2017learning} then extended~\cite{das2017visual} with an `image guessing' game, \ie, finding a desired image from a lineup of images through multi-round dialog. Reinforcement Learning (RL) was used to tackle this task. Later methods to visual dialog include applying Parallel Attention to discover the object through dialog~\cite{zhuang2018parallel}, learning a conditional variational auto-encoder for generating entire sequences of dialog~\cite{Massiceti_2018_CVPR}, unifying visual question generation and visual question answering in a dual learning framework~\cite{jain2018two}, combining RL
and Generative Adversarial Networks (GANs) to generate more human-like answers~\cite{Wu_2018_CVPR}.
In~\cite{jain2018two}, a discriminative visual dialog model was proposed and a new evaluation protocol was designed to test the questioner side of visual dialog. More recently, \cite{Kottur_2018_ECCV} used a neural module network to solve the problem of visual coreference resolution.

\noindent\textbf{Graph Neural Networks}~\cite{gori2005new,scarselli2009graph} draw a growing interest in the machine learning and computer vision communities, with the goal of combining structural representation of graph/graphical models with neural networks. There are two main stream of approaches. One stream is to design neural network operations to directly operate on graph-structured data~\cite{duvenaud2015convolutional,niepert2016learning,monti2016geometric,simonovsky2017dynamic,defferrard2016convolutional,kipf2017semi}. Another stream is to build graphically structured neural networks to approximate the learning/inference process of graphical models~\cite{lin2015deeply,sukhbaatar2016learning,li2016gated,fang2018learning,battaglia2016interaction,gilmer2017neural,wang2018attentive,chu2016crf}. Our method falls into this category. Some of these methods~\cite{lin2015deeply,sukhbaatar2016learning,battaglia2016interaction,gilmer2017neural,kipf2018neural,Qi_2018_ECCV} implement every graph node as a small neural network and formulate the interactions between nodes as a message propagation process, which is designed to be end-to-end trainable. Some others~\cite{zheng2015conditional,lin2015deeply,liu2015semantic,lin2016efficient,chu2016crf} tried to integrate CRFs and neural networks in a fully differentiable framework, which is quite meaningful for semantic segmentation. 

In this work, for the first time, we generalize the task of visual dialog to such a setting that we have partial observation of the nodes (\ie, image, caption and dialog history), and the graph structure (relations in dialog) needs to be automatically inferred. In this setting, the answer is the essentially unobserved node needs to be inferred based on the dialog graph, where the graph structure describes the dependencies among given dialog entities. We propose an essential neural network approach as an approximation to the EM solution of this problem. The proposed GNN is significantly different from most previous GNNs, which consider problems that the node features are observable, and usually a graph structure is given.

\vspace{-3pt}
\section{Our Approach}\label{sec:oa}
\vspace{-3pt}

We begin by describing the visual dialog task setup as introduced by Das \etal~\cite{das2017visual}. Formally, a visual dialog agent is given a dialog tuple $D\!=\!\{(I, C, H_t, Q_t)\}$ as input, including an image $I$, a caption $C$, a dialog history till round $t\!-\!1$, $H_t\!=\!\{(Q_k, A_k), k\!=\!1,\cdots,t\!-\!1\}$, and the current question $Q_t$ being asked at round $t$. The visual dialog agent is required to return a response $A_t$ to the question $Q_t$, by ranking a list of 100 candidate answers.

\begin{figure*}[t]
  \centering
      \includegraphics[width=1 \linewidth]{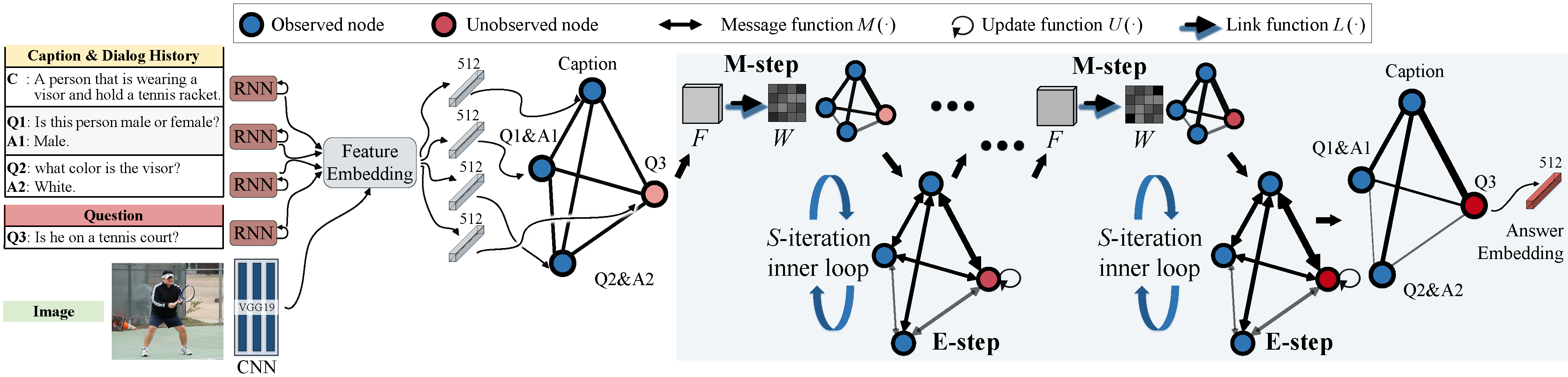}
\caption{\small The visual dialog is represented by a GNN, in which the dialog entities (\ie, caption, question \& answer pairs, and the unobserved queried answer) are represented by nodes (embeddings). The edges represent semantic dependencies between nodes.
Some nodes's values are observed (\ie, nodes that represent the dialog history), and we need to infer the missing values for the unobserved node (\ie, the queried answer) based on the underlying dialog structure.
The forward pass of the network emulates an EM algorithm, in which the M-step estimates the edge weights and E-step updates all hidden node states (embeddings) by neural message passing. After a few iterations, the hidden state for the unobserved node (answer) contains the inferred embedding for the missing value.}
\label{fig:model}
\vspace{-12pt}
\end{figure*}

In our approach, we represent the entire dialog by a graph, and we solve for the optimal queried answer by a GNN as an approximate inference (see \autoref{fig:model}). In this graph, the dialog entities $H_t\!=\!\{(Q_k, A_k), k\!=\!1,\cdots,t\!-\!1\}$, $Q_t$, and $A_t$ are represented as nodes. The graph structure (\ie, edges) represents the semantic dependencies between those nodes. The joint distribution of all the question and answer nodes are described by a Markov Random Field, where the values for some nodes are observed (\ie, the history questions \& answers, the current question). The node value for the current answer is unknown, and the model needs to infer its value as well as the graph structure encoded by the edge weights in this MRF.

The joint distribution in this MRF over all the nodes is specified by its potential functions and the graph structure. The potential functions can be learned in the training phase to maximize the likelihood of the training data, and used for inference in the testing phase. However, we cannot learn a fixed graph structure for all dialogs since they are different from dialog to dialog. For dialogs in both training and testing, we need to automatically infer the semantic structures.

In addition, because there is no label (also is hard to obtain) for such graph structures, our model needs to infer them in an unsupervised manner. Viewing the input nodes (\ie, the history questions \& answers, the current question) as observed data, the queried answer node as missing data, we adopt an EM algorithm to recover both the distribution parameter (the edge weights) and the missing data (the current answer). In this algorithm, the edge weights and the queried answer node are inferred to maximize the expected log likelihood. Finally, we resemble this inference process by a GNN approach, in which the node values and edge weights are updated in an iterative fashion.


\subsection{Dialog as Markov Random Field}
We model a dialog as an MRF, in which the nodes represent questions and answers and the edges encode semantic dependencies. Specifically, in a fully connected MRF model, the joint probability of all the nodes $\bm{v}$ is:
\vspace{-2pt}
\begin{equation}\small
\!\!p(\bm{v})\!=\!\frac{1}{Z}\exp{\{-\!\sum\nolimits_{i} \!\phi_{u}(v_i)\!-\!\sum\nolimits_{(i, j)\in E} \!\phi_p(v_i, v_j)\}},\!\!
\vspace{-2pt}
\end{equation}
where $Z$ is a normalizing constant, $\phi_{u}(v_i)$ is the unary potential function, and $\phi_p(v_i, v_j)$ is the pairwise potential function.

In our task, we want to learn a general potential function for all dialogs. We also want to maintain soft relations between nodes (\ie, a connectivity between 0 and 1) instead of just binary relations. Hence we generalize the above form to an MRF with $0 \sim 1$ weighed edges:
\vspace{-2pt}
\begin{equation}\small
\begin{aligned}
\!\!\!\!p(\bm{v} | W)&\!=\!\frac{1}{Z}\exp{\{-\!\sum\nolimits_{i} w_i \phi_{u}(v_i) - \!\sum\nolimits_{i, j} w_{ij} \phi_p(v_i, v_j)\}}\!\!\!\!\\
&\!=\! \frac{1}{Z}\exp{\{ - \!\Tr(W^T \Phi(\bm{v}))  \}},
\end{aligned}
\vspace{-2pt}
\end{equation}
where $w_i$ and $w_{i, j}$ are the weights that compose the edge weight matrix $W$. Note that we write $\Phi(\bm{v})$ the potential matrix as a compact form of all the potentials between nodes, where $\Phi_{i, i} = \phi_u(v_i)$ and $\Phi_{i, j} = \phi_p(v_i, v_j)$.

\subsection{Inference with Partial Observation}
Next we briefly review EM as a typical approach to do inference with missing data.
Suppose we have observed data $\bm{x}$ and unobserved data $\bm{z}$, whose joint distribution is parametrized by $\theta$. The goal is to infer the most likely parameter $\theta$ and random variable $\bm{z}$. The EM algorithm optimizes the expected log likelihood:
\vspace{-2pt}
\begin{equation}\small
\begin{aligned}
Q(\theta, \theta^{\text{old}}) = \int_z p(\bm{z} | \bm{x}, \theta^{\text{old}}) \log p(\bm{x}, \bm{z} | \theta) dz.
\end{aligned}
\label{eqn:expected_log_likelihood}
\vspace{-2pt}
\end{equation}

An EM algorithm is an iterative process of two steps: expectation (E-step) and maximization (M-step). In the E-step, the above expected likelihood is computed. In the M-step, the parameter $\theta$ is optimized to maximize this objective:
\begin{equation}\small
\begin{aligned}
\theta &= \argmax_{\theta} Q(\theta, \theta^{\text{old}}).
\end{aligned}
\end{equation}

The EM iteration always increases the observed data likelihood and terminates when a local minimum is found. However, the expected log likelihood \autoref{eqn:expected_log_likelihood} is often intractable. In the visual dialog task, to compute this quantity we need to enumerate all possible answers to the current question in the entire language space. In practice, we can use an surrogate objective in the E-step, in which we compute the plug-in approximation~\cite{van2000asymptotic} by a maximum a posteriori (MAP) estimate:
\vspace{-2pt}
\begin{equation}\small
\begin{aligned}
\tilde{Q}(\theta, \theta^{\text{old}}) &= \max_z p(\bm{z} | \bm{x}, \theta^{\text{old}}) \log p(\bm{x}, \bm{z} | \theta).
\end{aligned}
\label{eqn:Q_surrogate}
\vspace{-2pt}
\end{equation}
Then in the M-step we update the $\theta$ according to this surrogate objective.

\begin{figure*}[t]
  \centering
      \includegraphics[width=1 \linewidth]{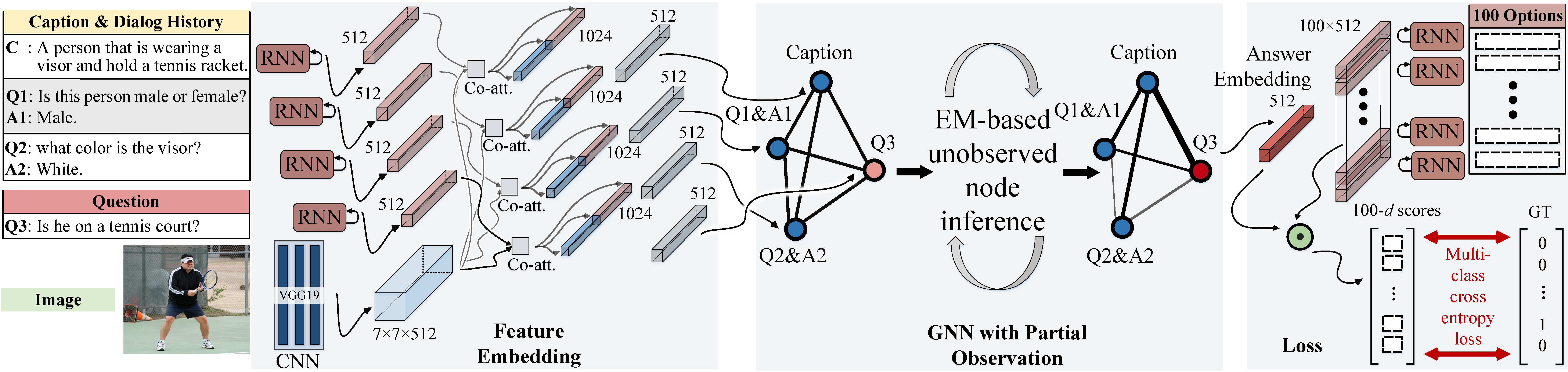}
\caption{\small A detailed illustration of our model. The left part shows feature extractions for each node, which serve as the initializations for node hidden states. After a few EM iterations, we obtain the hidden state (embedding) for the unobserved node (the queried answer). To choose the best answer from the pre-defined options, we use the dot product between the node and option embeddings as a similarity score. The scores are turned into probabilities by softmax activation, and a cross entropy loss is computed to train the network.}
\label{fig:details}
\vspace{-12pt}
\end{figure*}

\subsection{MRF with Partial Observations}
In the visual dialog task, the question \& answer history and the current question is given, hence we know the values for those nodes in the MRF. The task is to find out the missing value of the current answer node and the underlying sementic structure. Suppose in an MRF, we observe some nodes in the graph and we do not know the edge weights $W$. Denote the observed nodes as $\bm{x}$ and the unobserved nodes as $\bm{z}$, where $\bm{v} = \bm{x} \cup \bm{z}$ and $\bm{x} \cap \bm{z} = \emptyset$. Here the weight matrix $W$ parametrizes the joint distribution of $\bm{x}$ and $\bm{z}$, hence it can be viewed as the $\theta$ in the previous section. To jointly infer $W$ the graph structure (\eg, the semantic dependencies) and $\bm{z}$ the missing values (\eg, the queried answer), we run an EM algorithm:

\noindent\textbf{E-step:} We compute $\bm{z}^{*} = \argmax_z p(\bm{z} | \bm{x}, W^{\text{old}})$ to obtain $\tilde{Q}(\theta, \theta^{\text{old}})$ in \autoref{eqn:Q_surrogate}. This is achieved by a max-product loopy belief propagation~\cite{weiss2001optimality}. At every iteration, each node sends a (different) message to each of its neighbors and receives a message from each neighbor. After receiving message from neighbors, the belief $b(v_{i})$ for each node $v_{i}$ is updated by the max-product update rule:
\vspace{-2pt}
\begin{equation}\small
\begin{aligned}
b(v_{i}) = \alpha \phi_{u} (v_{i}) \prod\nolimits_{v_{j} \in \mathcal{N}(v_{i})} m_{ji}(v_i),
\end{aligned}
\vspace{-2pt}
\end{equation}
where $\alpha$ is a normalizing constant, $\mathcal{N}(v_{i})$ denotes the neighbor nodes of $v_{i}$, and $m_{ji}(v_i)$ is the message from $v_{j}$ to $v_{i}$. The message is given by:
\vspace*{-2pt}
\begin{equation}\small
\begin{aligned}
\!\!m_{ji}(v_i) = \max_{v_j} \, w_{ij} \, \phi_{p}(v_{i}, v_{j}) \!\prod\nolimits_{v_k \in \mathcal{N}(v_{j}) \setminus v_{i}} \!\!m_{kj}(v_{j}).
\end{aligned}
\vspace*{-4pt}
\end{equation}
where $\mathcal{N}(v_{j}) \setminus v_{i}$ indicates the all the neighboring nodes of $v_{j}$ except $v_{i}$.

\noindent\textbf{M-step:} Based on the estimated $\bm{z}^{*}$ in the E-step, we want to find the edge weights that maximizes the objective \autoref{eqn:Q_surrogate}:
\vspace*{-4pt}
\begin{equation}\small
\begin{aligned}
W &= \argmax_{W} \tilde{Q}(W, W^{\text{old}}) \\
&= \argmax_{W} p(\bm{z}^{*} | \bm{x}, W^{\text{old}}) \log p(\bm{x}, \bm{z}^{*} | W)\\
&= \argmax_{W} \log p(\bm{x}, \bm{z}^{*} | W).
\end{aligned}
\vspace*{-2pt}
\end{equation}

The M-step together with E-step forms a coordinate descent algorithm in the objective function $\tilde{Q}(W, W^{\text{old}})$. This algorithm contains two loops: an outer loop of inferring $\bm{z}$ and $\theta$ alternatively, and an inner loop of inferring the missing values $\bm{z}$ by iterative belief propagation.

Note that in the partial observed case, for the E-step we fix the observed nodes $v_{x}\!\in\!\bm{x}$ and only update the unobserved nodes $v_{z}\!\in\!\bm{z}$. Hence we also only need to compute messages from observed nodes to unobserved nodes. The message passing and belief update process iterate until convergence. When the iteration terminates, we obtain an MAP estimate $\bm{z}^{*}$ for the missing values, conditioned on the observed nodes $\bm{x}$ and current estimated edge weights $W$.



\subsection{GNN with Partial Observations}\label{sec:gnn}
We design a GNN for the visual dialog task guided by the above formulations. The network is structured as an MRF, in which the caption and each question/answer pair is represented as a node embedding, and the semantic relations are represented by edges. The model contains three different neural modules: message functions, update functions, and link functions. These modules are called iteratively to emulate the above EM algorithm.

\noindent\textbf{E-step:} We perform a neural message passing/belief propagation~\cite{gilmer2017neural} for an approximate inference of missing values $\bm{z}^{*}$. This process emulates the belief propagation in the E-step. For each node, we use an hidden state/embedding to represent its value. During belief propagation, the observed variables $\bm{x}$ and the edge weights $W$ are fixed. The hidden states of the unobserved nodes are iteratively updated by communicating with other nodes. Specially, we use message functions $M(\cdot)$ to summarize messages to nodes coming from other nodes, and update functions $U(\cdot)$ to update the hidden node states according to the incoming messages.

At each iteration step $s$, the update function computes a new hidden state for a node after receiving incoming messages:
\begin{equation}\small
\begin{aligned}
h_{v_{i}}^{s} = U(h_{v_{i}}^{s-1}, m_{v_{i}}^{s}),
\end{aligned}
\label{eqn:update_function}
\end{equation}
where $h_{v}^{s}$ is the hidden state/embedding for node $v$. $m_{v}^{s}$ is the summarized incoming message for node $v$ at $s$-th iteration. The messages are given by:
\vspace*{-3pt}
\begin{equation}\small
\begin{aligned}
m_{v_{i}}^{s} = \sum\nolimits_{v_{j} \in \mathcal{N}(v_{i})} w_{ij} M(h_{v_{i}}^{s-1}, h_{v_{j}}^{s-1}).
\end{aligned}
\label{eqn:message_function}
\end{equation}
The message passing phase runs for $S$ iterations towards convergence. At the first iteration, the node hidden states $h_{v}^{0}$ are initialized by node features $F_{v}$.

\noindent\textbf{M-step:} Based on the updated hidden states of all the nodes in the E-step, we update the edge weights $W$ by link functions. A link function $L(\cdot)$ estimates the connectivity $w_{ij}$ between two nodes $v_i$ and $v_j$ based on their current hidden states:
\begin{equation}\small
\begin{aligned}
w_{ij} = L(h_{v_i}, h_{v_j}).
\end{aligned}
\label{eqn:link_function}
\end{equation}



\subsection{Network Architecture}
\label{sec:arch}
At each round of the dialog, we aim to answer the query question based on the image, caption, and the question \& answer (QA) history.
For dialog round $t$, we construct $t\!+\!1$ nodes in which one node represents the caption, $t\!-\!1$ nodes represents the history of $t\!-\!1$ rounds of QAs, and one last node represents the answer to the current query question. The embedding for each node is initialized by fusing the image feature and the language embedding of the corresponding sentence(s). As shown in \autoref{fig:details}, for the caption node we extract the language embedding of the caption, and fuse it with the image feature as an initialization. For the last node representing the queried answer, we use the corresponding question embedding fused with the image feature to initialize the hidden state. For the rest nodes, the hidden states are initialized by fusing the QA embedding and the image feature. The fusing of language embeddings and image features are achieved by co-attention techniques~\cite{lu2016hierarchical}, and more details are introduced in \autoref{sec:ex}. The goal of our approach is to infer the hidden state of the queried answer by the emulated EM algorithm.

After initializing the node hidden states with feature embeddings, we start the iterative inference by first estimating the edge weights. The edge weights are estimated by \autoref{eqn:link_function}, where the link function is given by a dot product between transformed hidden states:
    \vspace{-4pt}
\begin{equation}\small
\begin{aligned}
w_{ij} = L(h_{v_i}, h_{v_j}) = \langle \text{fc}(h_{v_i}), \text{fc}(h_{v_j}) \rangle
\end{aligned}
\label{eqn:link_function_detail}
\vspace{-4pt}
\end{equation}
where $\langle \cdot,\cdot \rangle$ denotes dot product, and $\text{fc}(\cdot)$ denotes multiple fully connected layers with Rectified Linear Units (ReLU) in between the layers.

Using $M(h_{v_{i}}^{s-1}, h_{v_{j}}^{s-1}) = h_{v_{j}}^{s-1}$ as the message function, the summarized message from all neighbor nodes is computes as $m_{v_{i}}^{s} = \sum\nolimits_{v_{j} \in \mathcal{N}(v_{i})} w_{ij} h_{v_{j}}^{s-1}$. To stabilize the training of the update function, we normalize the sum of weights for edges coming into one node to 1 by a softmax function. Then the node hidden state is update by a Gated Recurrent Unit (GRU)~\cite{cho2014learning}:
\vspace{-4pt}
\begin{equation}\small
\begin{aligned}
h_{v_{i}}^{s} = U(h_{v_{i}}^{s-1}, m_{v_{i}}^{s}) = \text{GRU}(h_{v_{i}}^{s-1}, m_{v_{i}}^{s}).
\end{aligned}
\label{eqn:update_function}
\vspace{-4pt}
\end{equation}
Here the GRU is chosen for two reasons. First, \autoref{eqn:update_function} has a natural recurrent form. GRU is one type of Recurrent Neural Networks (RNN) that known to be more computationally efficient than Long short-term memory (LSTM). Second, Li \etal~\cite{li2016gated} has shown that GRU performs well in GNNs as update functions.

\begin{algorithm}[!t]\small
\caption{\small EM for Graph Neural Network}
\label{alg:alg}
\begin{flushleft}
\textbf{~~Input:} Extracted features $F_{v_x}$ for observed nodes $v_x \in \bm{x}$\\
\textbf{~~Output:} Graph structure $W$, node embeddings $h_{v_z}$ for unobserved nodes $v_z\!\in\!\bm{z}$
\end{flushleft}
\vspace{-10pt}
\begin{algorithmic}[1]
\STATE {/*~\textit{Initialization} */}
\FOR {each observed node $v_x\!\in\!\bm{x}$}
\STATE Initialize $h_{v_x}$ to be $F_{v_x}$
\ENDFOR
\FOR {each unobserved node $v_z\!\in\!\bm{z}$}
\STATE Initialize $h_{v_z}$ to be the question embedding
\ENDFOR
\STATE {/*~\textit{Expectation-Maximization: outer loop} */}
\WHILE {not converged}
\STATE {/*~\textit{M-step} */}
\FOR {each node pair $(v_{i}, v_{j})$}
\STATE $w_{ij} = L(h_{v_{i}}, h_{v_{j}}) = \langle \text{fc}(h_{v_i}), \text{fc}(h_{v_j}) \rangle$
\ENDFOR
\STATE {/*~\textit{E-step: inner loop for message passing} */}
\FOR {step $s$ from $1$ to $S$}
\FOR {each $v_{z}\!\in\!\bm{z}$}
\STATE {/*~\textit{Compute incoming message for $v_{i}$} */}
\STATE $m_{v_{z}}^{s} = \sum\nolimits_{v_{j}\in\mathcal{N}(v_{z})} w_{zj} h_{v_{j}}^{s-1}$
\STATE {/*~\textit{Update embedding for unobserved $v_{i}$} */}
\STATE $h_{v_{z}}^{s} = U(h_{v_{z}}^{s-1}, m_{v_{z}}^{s}) = \text{GRU}(h_{v_{z}}^{s-1}, m_{v_{z}}^{s})$
\ENDFOR
\ENDFOR
\ENDWHILE
\end{algorithmic}
\end{algorithm}

The algorithm stops after several iterations of the outer loop for EM, in which the edge weights $W$ and the node hidden states $h_v$ are updated alternatively. Inside each iteration, an inner loop is performed to update the node hidden states. The inner loop emulates the E-step, where a belief propagation is performed. The algorithm is illustrated in \autoref{alg:alg}. For the visual dialog task, the set of unobserved nodes include only the node that represents the current queried answer.

Finally, we regard the hidden state of the last node as the embedding of the queried answer. To choose one answer from the pre-defined options provided by the dataset, we compute $\langle h_v, h_o \rangle$ where $h_v$ is the node hidden state from the last node and $h_o$ is the language embedding for an option. A softmax activation function is applied to those dot products, and a multi-class cross entropy loss is computed to train the GNN.

\section{Experiments}\label{sec:ex}
\vspace{-3pt}
\subsection{Performance on VisDial v0.9~\cite{das2017visual}}\label{sec:vd}
\vspace{-3pt}
\noindent\textbf{Dataset:} We first evaluate the proposed approach on VisDial v0.9~\cite{das2017visual}, which was collected via two Amazon Mechanical Turk (AMT) subjects chatting about an image. The first person is allowed to see only the image caption, and instructed to ask questions about the hidden image to better understand the scene.  The second worker has access to both image and caption, and is asked to answer the first person's questions. Both are encouraged to talk in a natural manner. Their conversation is ended after 10 rounds of question answering. VisDial v0.9 contains a total of 1,232,870 dialog question-answer pairs on MSCOCO images~\cite{lin2014microsoft}. It is split into 80K for \verb"train", 3K  for \verb"val" and 40K as the \verb"test", in a manner consistent with~\cite{das2017visual}.

\noindent\textbf{Evaluation Protocol:} 
We follow~\cite{das2017visual} to evaluate individual responses at each round ($t\!=\!1,2,\cdots,10$) in a retrieval setup. Specifically, at test time, every question is coupled with a list of 100 candidate answer options, which a VisDial model is asked to return a sorting of the candidate answers. The model is
evaluated on standard retrieval metrics~\cite{das2017visual}:  Recall@1, Recall@5, Recall@10, Mean Reciprocal Rank (MRR), and
Mean Rank of human response. 
Lower value for MR and higher values for all the other metrics are desirable.

\noindent\textbf{Data Preparation:} To pre-process the data, we first resize each image into 224$\times$224 resolution, and use the output of the last pooling layer (\textit{pool5}) of VGG-19~\cite{simonyan2014very} as the image feature (512$\times$7$\times$7). For the text data, \ie, caption, questions and answers, we convert digits to words, and remove contractions, before tokenizing.
The captions, questions, answers longer than 40, 20, 20 words respectively are truncated.
All the texts in the experiment are lowercased. Each word is then turned into a vector representation with a look-up table, whose entries are 300-$d$ vectors learned along other parameters during training. Thus for caption, each question and answer, we have the sequences of word embedding with size of 40$\times$300, 20$\times$300, and 20$\times$300, respectively. The embedding of the caption, question or answer, is passed through a two-layered LSTM with 512 hidden states and the output state is used as our final text embeddings. We use the same LSTM and word embedding matrix across question, history, caption and options.

\noindent\textbf{Implementation Details:} We use 2 layers of fully connected layer in \autoref{eqn:link_function_detail}. The update function $U(\cdot)$ in \autoref{eqn:update_function} is implemented as a one-layer GRU with 512 hidden states. We use a single Titan Xp GPU to train the network with a batch size of 32. In the experiments, we use the Adam optimizer with a base learning rate of 1e-3 further decreasing to 5e-5. The training converges after $\sim$5 epochs.

\begin{table}[t!]
  \centering
  \resizebox{0.48\textwidth}{!}{
    \setlength\tabcolsep{4pt}
    \renewcommand\arraystretch{1}
  \begin{tabular}{r||ccccc}
  \hline\thickhline
  \rowcolor{mygray}
   Methods&MRR $\uparrow$ &\!R@1 $\uparrow$\! &\!R@5 $\uparrow$\! &\!R@10 $\uparrow$\! &\!Mean $\downarrow$\!\\
  \hline
  \hline
  LF~\cite{das2017visual}  &0.5807 &43.82 &74.68 &84.07 &5.78\\
  HRE~\cite{das2017visual} &0.5846 &44.67 &74.50 &84.22 &5.72\\
  HREA~\cite{das2017visual}  &0.5868 &44.82 &74.81 &84.36 &5.66\\
  MN~\cite{das2017visual}  &0.5965 &45.55 &76.22 &85.37 &5.46\\
  SAN-QI~\cite{yang2016stacked}  &0.5764 &43.44 &74.26 &83.72 &5.88\\
  HieCoAtt-QI~\cite{lu2016hierarchical}  &0.5788  &43.51  &74.49  &83.96  &5.84\\
  AMEM~\cite{seo2017visual}  &0.6160 &47.74 &78.04 &86.84 &4.99\\
  HCIAE-NP-ATT~\cite{lu2017best}  &0.6222 &48.48 &78.75 &87.59 &4.81\\
  SF~\cite{jain2018two} &0.6242 &48.55 &78.96 &87.75 &4.70\\
  SCA~\cite{Wu_2018_CVPR} &0.6398 &50.29 &80.71 &88.81 &4.47\\
  \hline
  \textbf{Ours}  &\textbf{0.6285}  &\textbf{48.95} & \textbf{79.65} & \textbf{88.36} & \textbf{4.57}\\
  \hline
  \end{tabular}
  }
  \vspace*{1pt}
  \caption{\textbf{Quantitative evaluation of discriminative methods  on
val split of VisDial v0.9~\cite{das2017visual}.} Our model outperforms most competitors.  See \autoref{sec:vd} for more details.
  }
  \vspace{-12pt}
  \label{tab:exqr}
\end{table}

\noindent\textbf{Quantitative Results:} We compare our method with several state-of-the-art discriminative dialog models, \ie, LF~\cite{das2017visual}, HRE~\cite{das2017visual}, HREA~\cite{das2017visual}, MN~\cite{das2017visual}, SAN-QI~\cite{yang2016stacked}, HieCoAtt-QI~\cite{lu2016hierarchical}, AMEM~\cite{seo2017visual}, HCIAE-NP-ATT~\cite{lu2017best}, SF~\cite{jain2018two}, and SCA~\cite{Wu_2018_CVPR}. \autoref{tab:exqr} summarizes the quantitative results of above competitors and our model. Our model consistently outperforms most approaches, highlighting the importance of understanding the dependencies in visual dialog. Specifically, our R@\textit{k}
($k$ = 1, 5, 10) is at least 0.4 point higher than SF. Our method only performs slightly worse than SCA, which adopts adversarial learning techniques.

\begin{figure*}[ht]
  \centering
      \includegraphics[width=1\linewidth]{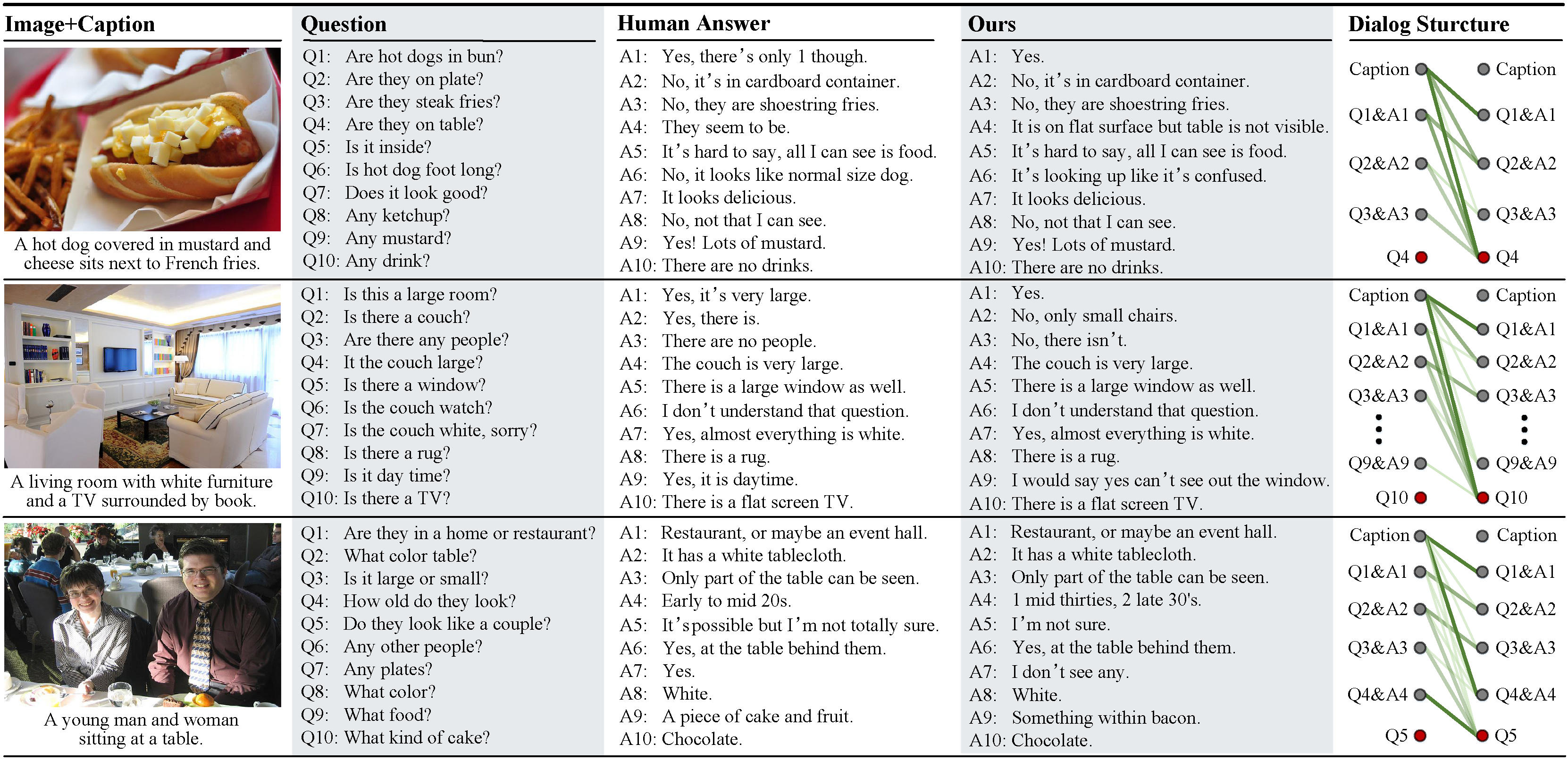}
\caption{\small Qualitative results of our model on VisDial v0.9~\cite{das2017visual}, comparing to human ground-truth answer. The last column presents the visual dialog structures inferred by our model, where the more darker green links indicate higher relations (predicted by link functions). }
\label{fig:qualitative}
\vspace{-12pt}
\end{figure*}

\noindent\textbf{Qualitative Results:} \autoref{fig:qualitative} shows some qualitative results of our model. We summarize three key observations: \textbf{(i)} We compare our machine selected answer with human answer and show that our model is capable of selecting meaningful yet different answers compared with the ground-truth answer.
\textbf{(ii)} We present our inferred dialog structure according to the edge weight between each pair of nodes. We show that the edge weight is relatively high when the correlation between the node pairs is strong.
\textbf{(iii)} \autoref{tab:exqr} and \autoref{fig:qualitative}
illustrate the interpretable and grounded nature of our model. As seen, the suggested model successfully captures the relations in dialog and attend to dialog fragments which are relevant to current question.

\vspace{-3pt}
\subsection{Performance on VisDial v1.0~\cite{das2017visual}}\label{sec:vdv1}
\vspace{-1pt}
\noindent\textbf{Dataset:} Then we test our model on the newest version of VisDial dataset~\cite{das2017visual}: VisDial v1.0, which is collected in a similar way of VisDial v0.9. For VisDial v1.0, all the VisDial v0.9 (\ie, 1,232,870 dialog question-answer pairs on MSCOCO images~\cite{lin2014microsoft}) is used for \verb"train", extra 20,640 and 8,000 dialog question-answer pairs are used for \verb"val" and \verb"test", respectively.

\noindent\textbf{Evaluation Protocol:} In addition to the five evaluation metrics (\ie, Recall@1, Recall@5, Recall@10, MRR, and Mean Rank of human response) used in VisDial v0.9, an extra metric, Normalized Discounted Cumulative Gain (NDCG), is involved for a more comprehensive quantitative performance study. Higher value for NDCG is better.

\noindent\textbf{Quantitative Results:} Five discriminative dialog models (\ie, LF~\cite{das2017visual}, HRE~\cite{das2017visual}, MN~\cite{das2017visual}, LF-Att~\cite{das2017visual}, MN-Att~\cite{das2017visual}) were included in our experiments. \autoref{tab:exvdv1} presents the overall quantitative comparison results. As seen, the suggested model consistently gaining promising results.

\begin{table}[t!]
  \centering
  \resizebox{0.49\textwidth}{!}{
    \setlength\tabcolsep{4pt}
    \renewcommand\arraystretch{1}
  \begin{tabular}{r||cccccc}
  \hline\thickhline
  \rowcolor{mygray}
   Methods&MRR $\uparrow$ &\!R@1 $\uparrow$\! &\!R@5 $\uparrow$\! &\!R@10 $\uparrow$\! &\!Mean $\downarrow$\! &NDCG $\uparrow$ \\
  \hline
  \hline
  LF~\cite{das2017visual} &0.5542 &40.95 &72.45 &82.83 &5.95 &0.4531 \\
  HRE~\cite{das2017visual} &0.5416 &39.93 &70.45 &81.50 &6.41 &0.4546 \\
  MN~\cite{das2017visual}  &0.5549 &40.98 &72.30 &83.30 &5.92 &0.4750 \\
  LF-Att~\cite{das2017visual} &0.5707 &42.08 &74.82 &85.05 &5.41 &0.4976 \\
  MN-Att~\cite{das2017visual} &0.5690 &42.42 &74.00 &84.35 &5.59&0.4958 \\
  \hline
  \textbf{Ours} &\textbf{0.6137}  &\textbf{47.33} &\textbf{77.98} &\textbf{87.83} &\textbf{4.57} &\textbf{0.5282}\\
  \hline
  \end{tabular}
  }
  \vspace*{2pt}
  \caption{\textbf{Quantitative evaluation of discriminative methods on test-standard split of VisDial v1.0~\cite{das2017visual}.} Our model outperforms all other models across all metrics. See \autoref{sec:vdv1} for more details.
  }
\vspace*{-12pt}
  \label{tab:exvdv1}
\end{table}

\vspace{-3pt}
\subsection{Performance on VisDial-Q Dataset~\cite{das2017visual,jain2018two}} \label{sec:vq}
\vspace{-1pt}
\noindent\textbf{Dataset:} VisDial Dataset~\cite{das2017visual} provides a solid foundation for assessing the performance of a visual dialog system answering questions. To test the questioner side of visual dialog, Jain \etal~\cite{jain2018two} further propose a VisDial-Q dataset, which is built upon VisDial v0.9~\cite{das2017visual}. The dataset splitting is the same as VisDial v0.9.

\noindent\textbf{Evaluation Protocol:} VisDial-Q dataset is companied with a retrieval based `\textit{VisDial-Q evaluation protocol}', analogous to the `\textit{VisDial evaluation protocol}' in VisDial dataset detailed before. A visual dialog system is required to choose one out of 100 next questions for a given question-answer pair. Similar methodology in~\cite{das2017visual} is adopted to collect the 100 follow-up question candidates.
Therefore, the metrics described in \autoref{sec:vd}: Recall@$k$, MRR, and Mean Rank, are also used for quantitative evaluation.

\noindent\textbf{Data Preparation:} We use the same text embedding techniques as we used for \autoref{sec:vd}. Different from VisDial task, the first round of QA pair is given to predict next round of question. Thus the maximum round of dialog in the VisDial-Q task is set as 9. Similar as we illustrate in \autoref{sec:arch}, we construct $t\!+\!1$ node with caption and previous history as the first $t$ nodes and the expected question as the last node. We initialize our question node with language embedding of the caption and set the language embedding of corresponding sentence as the embedding of the rest of nodes.


\begin{table}[t!]
  \centering
  \resizebox{0.46\textwidth}{!}{
    \setlength\tabcolsep{4pt}
    \renewcommand\arraystretch{1}
  \begin{tabular}{r||ccccc}
  \hline\thickhline
  \rowcolor{mygray}
  Methods&MRR $\uparrow$ &\!R@1 $\uparrow$\! &\!R@5 $\uparrow$\! &\!R@10 $\uparrow$\! &\!Mean $\downarrow$\!\\
  \hline
  \hline
  SF-QI~\cite{jain2018two} & 0.3021 & 17.38 & 42.32 & 57.16 & 14.03 \\
  SF-QIH~\cite{jain2018two} &0.4060 &26.76 &55.17 &70.39 &9.32\\
  \hline
  Ours (\textit{w/o iter}) & 0.3977 & 25.69 & 54.52 & 70.33 & 9.38 \\
  Ours (\textit{const. graph}) & 0.4025 & 26.08 & 55.30 & 70.83 & 9.24 \\
  \hline
    \textbf{Ours (\textit{full, 3 iter})}  &\textbf{0.4126}  &\textbf{27.15} & \textbf{56.47} & \textbf{71.97} & \textbf{8.86}\\
  \hline
  \end{tabular}
  }
  \vspace*{2pt}
  \caption{\textbf{Quantitative evaluation on VisDial-Q dataset~\cite{das2017visual,jain2018two}} with VisDial-Q evaluation protocol. See \autoref{sec:vq} for more details.
  }
     \vspace{-12pt}
  \label{tab:visdial_q}
\end{table}

\noindent\textbf{Quantitative Results:}
We follow the same protocol described in~\cite{jain2018two} to evaluate our model. \autoref{tab:visdial_q} shows the quantitative results for comparative methods and our ablative model variants. The ablative models include i) our model with constant graph (all edge weights are 1), and ii) our model without the EM iterations. Our full model with 3 EM iterations outperforms the comparative method in all evaluation metrics. Particularly, we can see that our model with constant graph has a similar performance to the comparative method. This shows the effectiveness of our EM-based inference process. Experiment results on this dataset also shows the generality of our approach: it can infer the underlying dialog structure and reason accordingly about unobserved nodes (next question or current answer).


\vspace{-3pt}
\subsection{Diagnostic Experiments}\label{sec:as}
\vspace{-1pt}
To assess the effect of some essential component of our model, we implement and test several variants: \textbf{(i)} constant graph that fixes edge weight between each pair of nodes to be 1; \textbf{(ii)} graph without EM iteration; and \textbf{(iii)} graph with \textit{n} EM iterations. \autoref{tab:ablative} shows the quantitative evaluations of these model variants on VisDial v0.9~\cite{das2017visual}. We summarize our observations here: \textbf{(a)} model without EM iterations performs the worst among all variants. This shows the importance of iteratively updating the node embeddings. \textbf{(b)} In our experiments, message passing with 3 iterations shows the best performance of our proposed model. \textbf{(c)} model using constant graph (3 iterations) performs better than worse than the model without EM iterations, since it allows iterative updates of node embeddings. However, it is outperformed by other iterative models with a dynamic structure, since all incoming messages are treated equally. This shows the importance of edge weights: they filter out misleading messages while allowing information flow.

\begin{table}[t!]
  \centering
  \resizebox{0.45\textwidth}{!}{
    \setlength\tabcolsep{6pt}
    \renewcommand\arraystretch{1}
  \begin{tabular}{r||ccccc}
  \hline\thickhline
  \rowcolor{mygray}
   Methods&MRR $\uparrow$ &\!R@1 $\uparrow$\! &\!R@5 $\uparrow$\! &\!R@10 $\uparrow$\! &\!Mean $\downarrow$\!\\
  \hline
  \hline
  \textbf{Ours \textit{(3 iter)}}.  &\textbf{0.6285}  &\textbf{48.95} & \textbf{79.65} & \textbf{88.36} & \textbf{4.57}\\
    \hline
    \textit{const. graph}. & 0.6197 & 47.91 & 78.99 & 87.77 & 4.74\\
    \textit{w/o iter}. & 0.6162 & 46.73 & 78.41 & 87.26 & 4.84\\
    \textit{2 iter}. & 0.6213 & 48.18 & 78.97 & 87.81 & 4.75 \\
    \textit{4 iter}. & 0.6237 & 48.41 & 79.20 & 87.95 & 4.68 \\
  \hline
  \end{tabular}
  }
   \vspace*{2pt}
  \caption{\textbf{Ablation study of the key components of our methods on VisDial v0.9 dataset~\cite{das2017visual}.} See \autoref{sec:as} for more details.
  }
    \vspace{-12pt}
  \label{tab:ablative}
\end{table}

\vspace{-3pt}
\section{Conclusion}\label{sec:conc}
\vspace{-3pt}
In this paper, we develop a novel model for the visual dialog task. The backbone of this model is a GNN, in which each node represents a dialog entity and the edge weights represent the semantic dependencies between nodes. An EM-style inference algorithm is proposed for this GNN to estimate the latent relations between nodes and the missing values of unobserved nodes. Experiments are performed on the VisDial and VisDial-Q dataset. Results show that our method is able to find and utilize underlying dialog structures for dialog inference in both tasks, demonstrating the generality and effectiveness of our method.

\noindent\textbf{Acknowledgements} We thank Prof. Ying Nian Wu from UCLA Statistics Department for helpful discussions. This work reported herein was supported by DARPA XAI grant N66001-17-2-4029, ONR MURI grant N00014-16-1-2007 and ARO grant W911NF-18-1-0296.

{\small
\bibliographystyle{ieee_fullname}
\bibliography{egbib}
}

\end{document}